%% file: main.tex
 \documentclass[final,1p,times,nopreprintline]{elsarticle}

\usepackage{url}
\cortext[mycorrespondingauthor]{Corresponding author. 
Email address: \url{k-shimizu@rs.tus.ac.jp}~(K. Shimizu).}
\usepackage{amssymb}

\usepackage{amsmath}
\usepackage{bm} 
\usepackage{graphicx}
\usepackage{wrapfig} 
\usepackage{makeidx} 
\usepackage{here}
\usepackage{subcaption} 
\usepackage{algorithmic}
\usepackage{algorithm}
\usepackage{booktabs}
\usepackage{multirow}
\usepackage{tabularx} 
\usepackage{ulem} 
\usepackage[T1]{fontenc}
\usepackage{xcolor}
\usepackage{url,hyperref,lineno,microtype,subcaption}

\newtheorem{thm}{Theorem}[section]
\newtheorem{prop}[thm]{Proposition}
\renewcommand{\vec}[1]{\bm{#1}} 
\usepackage{array} 
\newcolumntype{L}[1]{>{\hspace{#1}}l} 

\journal{Annals of Mathematics and Artificial Intelligence}

\begin{document}

\begin{frontmatter}


\title{Evaluating Singular Value Thresholds \\for DNN Weight Matrices based on Random Matrix Theory} 


\author[a]{Kohei Nishikawa}
\author[a]{Koki Shimizu}
\author[a]{Hiroki Hashiguchi}

\address[a]{Tokyo University of Science,
            1-3 Kagurazaka, 
            Shinjuku-ku,
            162-8601, 
            Tokyo,
            Japan}

\begin{abstract}
This study evaluates thresholds for removing singular values from singular value decomposition-based low-rank approximations of deep neural network weight matrices. Each weight matrix is modeled as the sum of signal and noise matrices. The low-rank approximation is obtained by removing noise-related singular values using a threshold based on random matrix theory. 
To assess the adequacy of this threshold, we propose an evaluation metric based on the cosine similarity between the singular vectors of the signal and original weight matrices. The proposed metric is used in numerical experiments to compare two threshold estimation methods. 

\end{abstract}


\begin{keyword} 
Deep Learning \sep
Denoising \sep
Marchenko--Pastur distribution \sep
Random matrix 
\MSC[2010]  
 60B20 \sep
 62H10 \sep
 68T05
\end{keyword}

\end{frontmatter}



\input{section1}

\input{section2}
\input{section3}
\input{section4}
\input{section5}

\section*{Acknowlegments}
This work was supported by JSPS KAKENHI Grant Numbers 23K11016 and 25K17300.
\input{appendix}





\makeatletter
\renewcommand{\@biblabel}[1]{#1.}
\makeatother
\bibliographystyle{elsarticle-num-names} 
\bibliography{refs}

\end{document}

%% file: section1.tex
\section{Introduction}
\label{sec1}
Deep neural networks (DNNs) have been widely used in fields such as image processing, speech recognition, and natural language processing. 
However, their over-parameterized architectures tend to overfit the training data, which may lead to degraded generalization performance on unseen data \citep{zhang2021understanding,arpit2017closer}.
Various regularization techniques, such as weight decay \citep{krogh1991simple}, dropout \citep{srivastava2014dropout}, and network pruning \citep{han2015learning} have been proposed to reduce overfitting. Although these methods are effective in practice, many are designed and applied based on empirical heuristics.
Random matrix theory (RMT) has recently attracted attention as an approach that mitigates overfitting. 
The elements of a matrix are treated as random variables in RMT; moreover, it utilizes eigenvalue and singular value distributions to understand phenomena across various fields. 
In particular, the universal laws of random matrices enable the distinction between noise and signals in data and support noise reduction in a wide range of fields, including acoustic signal processing \citep{lu08}, single-cell technology \citep{aparicio20}, and financial correlation analysis \citep{plerou02}. 

Recently, RMT has also been applied to DNNs, spectral analysis of weight matrices \citep{thamm22,martin21a,martin21b}, early stopping criteria \citep{Meng23}, analysis of the statistical properties of the Hessian \citep{baskerville22}, and detection of grokking phenomena \citep{prakash25}.
Staats et al. \cite{staats23} reported that singular values of the weight matrices that follow the Marchenko--Pastur (MP) distribution may reflect less essential or redundant features for the task, whereas a few large singular values deviate from it.
They demonstrated that removing the small singular values has minimal impact on prediction accuracy, while 
yielding low-rank weight matrices that reduce redundant parameters and overall model complexity.
Building on this concept, Berlyand et al. \cite{berlyand24} proposed an RMT-based low-rank approximation method that removes singular values below a theoretically derived threshold.
However, various methods exist for determining such thresholds, rendering quantitatively evaluating the most appropriate method important.

In this paper, we present an evaluation metric based on RMT to assess singular value thresholds for separating signals from noise in DNN weight matrices. 
In Section \ref{sec2}, the relationship between RMT and DNN is discussed. 
The weight matrix is modeled as a perturbed matrix composed of a signal matrix that retains predictive information and a random matrix that does not. 
In Section \ref{sec3}, a similarity measure is proposed for the signal and low-rank approximated weight matrices, using the inner product of their respective singular vectors based on the theoretical framework of Benaych--Georges and Nadakuditi \cite{georges12}. 
In Section \ref{sec4}, the presented similarity metric is applied to the weight matrices of convolutional neural networks (CNNs), to evaluate whether the thresholding method of Ke et al. \cite{zheng21} or Gaussian broadening is more appropriate.

%% file: section2.tex
\section{Fitting the MP Distribution to the Singular Value Distribution of DNN Weight Matrices}
\label{sec2}
Let \(\vec{x}_i\) be the input data and \(\vec{y}_i\) the output data. DNNs with $L$ layers are represented using the number of nodes \(n_l\) in the \(l\)-th layer \((1 \leq l \leq L)\), activation function \(h_l(\cdot)\), weight matrix \(W_l \in \mathbb{R}^{n_l\times n_{l-1}} \), and bias vector \(\bm{b}_l\) as follows:
\[
E_{\rm DNN}(\vec{x}_i) = h_L \left( h_{L-1}\left( h_{L-2}\left( \cdots \right) W_{L-1} + \bm{b}_{L-1} \right) W_L + \bm{b}_L \right).
\]

The weight matrix \(W_l\) is determined by minimizing the loss \(\mathcal{L}\) between \(E_{\text{DNN}}(\vec{x}_i)\) and \(\vec{y}_i\) as follows:

\[
\min_{W_l, \bm{b}_l} \left( \sum_i \mathcal{L} \left(E_{\text{DNN}}(\vec{x}_i), \vec{y}_i \right) + \lambda \lVert  W_{l}\rVert \right),
\]
where \(\lVert \cdot \rVert\) denotes an arbitrary matrix norm and \(\lambda\) is a regularization parameter.
Each entry of the weight matrix is typically initialized randomly using distributions such as the Glorot uniform distribution \citep{glorot10}. The training process relies on optimization algorithms, such as the stochastic gradient descent (SGD) and its variants, requiring the careful tuning of hyperparameters (e.g., batch size and learning rate) for effective learning.
Hereafter, we simply denote the weight matrices in the \(l\)-th layer of a DNN as \(W \in \mathbb{R}^{n \times m}~(n \geq m)\).

The trained weight matrix $W \in \mathbb{R}^{n \times m}$ was modeled by Staats et al. \cite{staats23} as the sum of a signal matrix \(W_{\rm signal}\) and a random matrix \(W_{\rm noise}\), given by
\begin{align}
W = W_{\rm signal} + W_{\rm noise},  \label{perturbation}
\end{align}
where the entries of \(W_{\rm noise}\) are assumed to be independent and identically distributed (i.i.d.) with zero mean and variance $\sigma^2<\infty$.
The weight matrix is randomly initialized before training. As training progresses, signal components gradually emerge. 
The perturbation model in \eqref{perturbation} is commonly used in the analysis of DNN weight matrices based on RMT \citep{berlyand24,berlyand25}.
In the Appendix of Staats et al. \cite{staats23}, the matrix \(W_{\rm noise}\) is regarded as a random matrix with i.i.d. entries when the weights are optimized using SGD.

Next, we introduce the MP distribution \citep{marcenko67}, which is useful for removing redundant information from the weight matrices of the trained DNNs.
If \(n, m \to \infty\) with \( \frac{m}{n} \to q \in (0, 1] \), the singular values of \(W_{\rm noise}\) are known to follow the MP distribution, with density given by
\begin{equation}
g(x) = \frac{1}{\pi q \sigma^2 x} \sqrt{(x^2 - x_{\rm min}^2)(x_{\rm max}^2 - x^2)} \label{mar}, \quad x_{\rm min} \leq x \leq x_{\rm max},
\end{equation}
where \(x_{\rm max} = \sigma(1 + \sqrt{q})\) and \(x_{\rm min} = \sigma(1 - \sqrt{q})\).
The convergence rate of the spectral distribution is $O(n^{-1/4})$ if the ratio $q$ is far from $1$, and $O(n^{-5/48})$ if it is close to $1$. For further developments, see Bai and Silverstein \cite{bai2010spectral} and the references therein.
The MP distribution has a scaling parameter $\sigma$, which can be estimated using the bulk eigenvalue matching analysis (BEMA) or the Gaussian broadening approach.
For details on these estimation methods, see Appendices \ref{app1} and \ref{app2}.
Figure~\ref{fig:mp_estimation} shows the estimated MP distributions from $W$ of a multilayer perceptron (MLP) trained on MNIST dataset. 
The red and blue curves indicate the MP distributions estimated by BEMA and Gaussian broadening, respectively, with the corresponding vertical lines representing the noise--information boundaries.
\begin{figure}[H]
\centering
\includegraphics[width=0.6\textwidth]{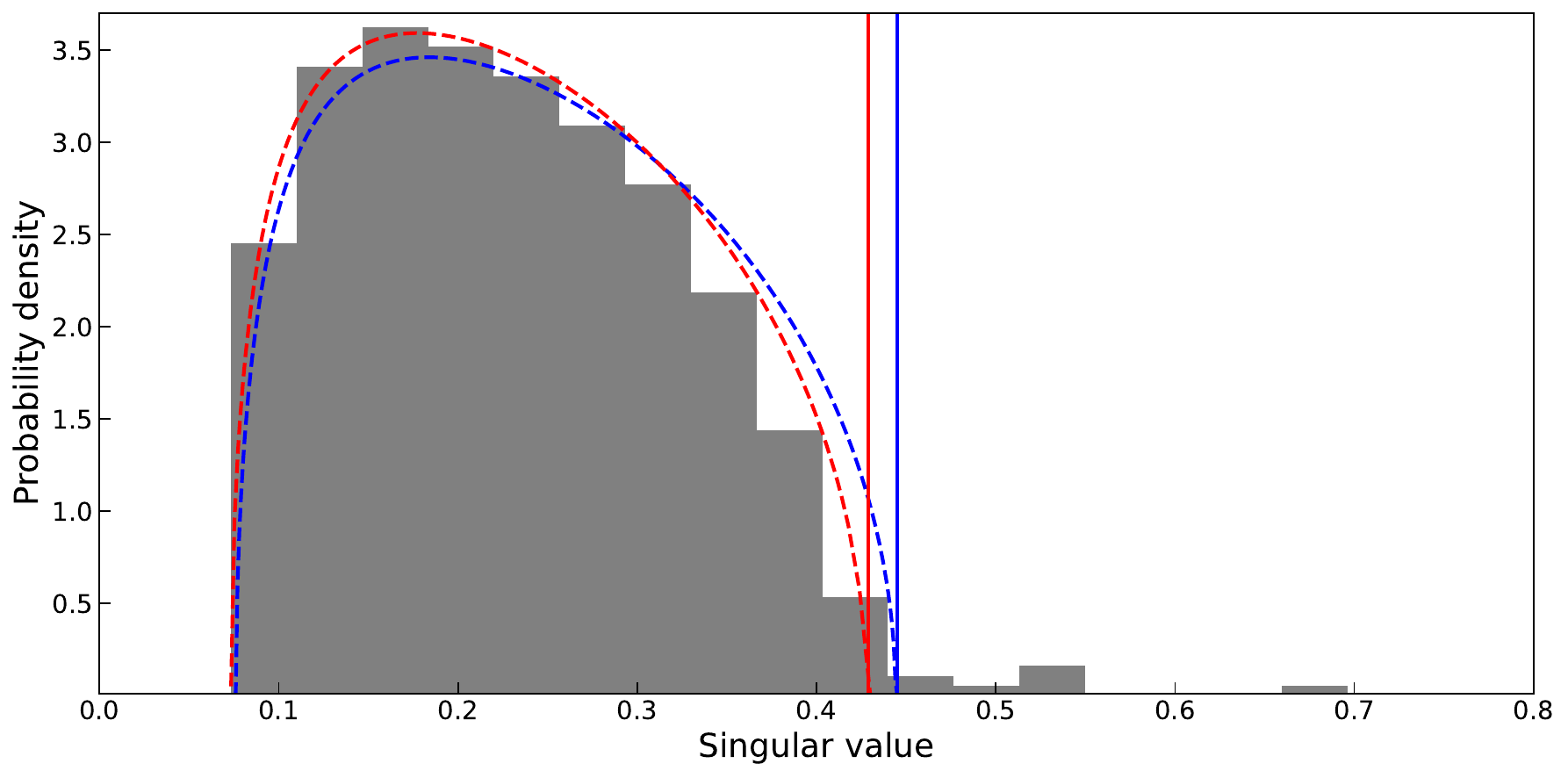}
\caption{Histogram of the singular values of $W$ in the MLP and density curve of the estimated MP distribution. The vertical red and blue lines indicate the thresholds estimated by BEMA and Gaussian broadening, respectively.}
\label{fig:mp_estimation}
\end{figure}
The singular values of \(W\) that fall within the support of the MP distribution are regarded as noise. 
In contrast, the singular values outside the support are interpreted as components derived from the signal matrix.
It is reported that the singular value distributions of trained weight matrices can, in some cases, be approximated by the MP distribution, extending beyond SGD training.
The fitting of empirical singular value distributions to the MP distribution in Transformer-based models is investigated in Dantas et al. \cite{dantas2025decoding} and Staats et al. \cite{staats25}.
\cite{staats23} estimated the MP distribution using BEMA,
whereas Berlyand et at. \cite{berlyand24} estimated it using Gaussian broadening.
They then used these estimates to perform low-rank approximation of weight matrices. However, there are discrepancies in the thresholds estimated by the BEMA and Gaussian broadening methods. This study aims to evaluate which estimation provides the more appropriate threshold.

%% file: section3.tex
\section{Metric for Evaluating Singular Value Thresholds}
\label{sec3}
In this section, we propose an evaluation metric to assess the threshold that distinguishes the singular values attributed to the signal matrix from those attributed to noise.
The singular value decomposition (SVD) of matrices \(W_\mathrm{signal}\) and \(W\) are given by
\begin{align*}
W_\mathrm{signal} = \sum_{i=1}^s \theta_i \bm{u}_i \bm{v}_i^{\top}, \quad 
W = \sum_{i=1}^m \gamma_i \tilde{\bm{u}}_i \tilde{\bm{v}}_i^{\top},
\end{align*}
where $\theta_1 \geq \theta_2 \geq \cdots \geq \theta_s\) and \(\gamma_1 \geq \gamma_2 \geq \cdots \geq \gamma_m$ are the singular values of $W_{\rm signal}$ and $W$, respectively.
The corresponding left and right singular vectors are denoted by \(\bm{u}_i, \tilde{\bm{u}}_i\) and \(\bm{v}_i, \tilde{\bm{v}}_i\), respectively.
The unknown parameter $s<m$ represents the number of singular values exceeding $\gamma_+$, which is given by
\[
s = \#\left\{ 1 \leq k \leq m : \gamma_k^2 >\gamma_+^2 \right\}.
\]
The upper threshold $\gamma_+ > 0$ of the MP distribution in Ke et al. \cite{zheng21} is given by
\begin{align}
  \gamma_+^2= \sigma^2 \left[(1 + \sqrt{q})^2 + t_{1 - \beta} \, n^{-2/3} q^{-1/6} (1 + \sqrt{q})^{4/3} \right], \label{threshold}
\end{align}
where $t_{1-\beta}$ is the upper $\beta$ percentile point of the Tracy--Widom (TW) distribution \citep{johnstone01}, with $\beta \in (0,1)$ being a hyperparameter.
The first term on the right-hand side of \eqref{threshold} represents the theoretical upper bound \(x_{\text{max}}\) of the MP distribution. 
In an asymptotic framework, the optimal hard threshold was proposed by Gavish and Donoho \cite{gavish2014optimal}.
However, in finite-size settings, random components may be mistakenly identified as signals, potentially leading to an overestimation of the number of signal components.  
Therefore, a correction term based on the TW distribution is considered, as it characterizes the distribution of the largest eigenvalue in RMT.

For the parameter $s$, the low-rank approximation for $W$ is given by
\[
\label{eq:W_LR}
W_\mathrm{LR} = \sum_{i=1}^s \gamma_i \tilde{\bm{u}}_i \tilde{\bm{v}}_i^{\top}.
\]
The low-rank approximation $W_{\mathrm{LR}}$ can be represented as the product of two matrices of dimensions $n \times s$ and $s \times m$.
If $s < nm/(n + m)$, the number of parameters is reduced from the original $nm$ to $s(n + m)$.
This decomposition enables model compression while preserving predictive performance.
As convolutional layer weights in CNN are fourth-order tensors, Zhang et al. \cite{zhang16} used a reshape-based method that converts the tensor into a matrix before applying a low-rank approximation.
The following proposition quantifies how well $W_{\mathrm{LR}}$ approximates $W_{\mathrm{signal}}$.
\begin{prop}[Benaych-Georges and Nadakuditi, 2012]
If \( n, m \to \infty\) and \(\frac{m}{n} \to q \in (0,1]\), the singular values \(\theta_i\) and the squared cosine similarity $\phi_i$ between singular vectors \(\tilde{\mathbf{u}}_i\) and \(\mathbf{u}_i\) satisfy
\begin{align*}
\theta_i \overset{\rm a.s.}{\to}
\frac{\sigma}{\sqrt{2}} \sqrt{\left(\frac{\gamma_i}{\sigma}\right)^2 - q - 1 + \sqrt{\left(\left(\frac{\gamma_i}{\sigma}\right)^2 - q - 1\right)^2 - 4q}},
\end{align*}
\begin{align}
\label{phi}
\phi_i =  |\langle \tilde{\bm{u}}_i, \bm{u}_i \rangle |^2 \overset{\rm a.s.}{\to} 
\frac{-2h(\rho_i)}{\theta_i^2 D'(\rho_i)},
\quad \theta_i \geq \sigma \cdot q^{1/4},
\end{align}
almost surely, where
\begin{align*}
D(z) &= \frac{z^2 - \sigma^2(q+1) - \sqrt{(z^2 - \sigma^2(q+1))^2 - 4\sigma^4 q}}{2\sigma^4 q}, \\
h(z) &= \int \frac{z}{z^2 - t^2}\, dg(t), \quad 
\rho_i = D^{-1}\left(\frac{1}{\theta_i^2}\right).
\end{align*}
The symbol \(D'\) denotes the derivative of \(D\), and \(g(t)\) is the probability density function of the MP distribution given in \eqref{mar}.
\end{prop}
If \(\sigma = 1\) in \eqref{phi}, the explicit expression is given by
\begin{align*}
\phi_i \overset{\rm a.s.}{\to} 1 - \frac{q(1 + \theta_i^2)}{\theta_i^2(\theta_i^2 + q)}.
\end{align*}
However, for general \(\sigma\), no closed-form expression is known, and a numerical evaluation of \(\phi_i\) is required.

We propose employing the cosine similarity \(\phi_i\) as an evaluation metric for assessing the similarity between the low-rank and signal matrices, defining the weighted average similarity by
\begin{align}
{\rm Ave}_w(\phi) = \frac{\sum_{i=1}^{s} \phi_i (\gamma_i - \gamma_+)}{\sum_{i=1}^{s}(\gamma_i - \gamma_+)}.
\label{avephi}
\end{align}
The similarity ${\rm Ave}_w(\phi)$ takes values within the interval $[0,1]$, where larger values of ${\rm Ave}_w(\phi)$ indicate that $W_{\text{LR}}$ is closer to $W_{\text{signal}}$.
The simple average of $\phi_i$ is an alternative metric to $\eqref{avephi}$.
However, if only the first few signal singular values are large while the rest lie near the bulk, the metric can become small even when the accuracy is maintained, leading to a loss of correspondence between the metric and the accuracy.
To facilitate better consistency with the accuracy, we employ the metric as weighted average.
Computing ${\rm Ave}_w(\phi)$ requires estimating the unknown parameter \(\sigma^2\). 
The parameter $\sigma^2$ can be estimated by BEMA or Gaussian broadening, to obtain $\hat{s}$.
Thus, the metric $\mathrm{Ave}_{w}(\phi)$ can be estimated through the following steps:
\begin{enumerate}
  \item Estimate the parameters $\sigma^2$ and $s$, as $\hat{\sigma}$ and $\hat{s}$, respectively.
  \item Estimate the singular values $\theta_i$ in \eqref{phi}\ for $i = 1, \dots, \hat{s}$ as

  \begin{align*}
\hat{\theta_i} =
\frac{\hat{\sigma}}{\sqrt{2}} \sqrt{\left(\frac{\gamma_i}{\hat{\sigma}}\right)^2 - q - 1 + \sqrt{\left(\left(\frac{\gamma_i}{\hat{\sigma}}\right)^2 - q - 1\right)^2 - 4q}}.
\end{align*}

  \item Estimate the cosine similarities $\phi_i$ in \eqref{phi}\ for $i = 1, \dots, \hat{s}$ as 
  
  \begin{align*}
\hat{\phi}_i =  
\frac{-2h(\hat{\rho}_i)}{\theta_i^2 D'(\hat{\rho}_i)}
\quad \text{with} \quad \hat{\rho}_i = D^{-1}\left(\frac{1}{\hat{\theta}_i^2}\right).
\end{align*}

  \item Compute the estimate of ${\rm Ave}_w(\phi)$ in \eqref{avephi} as
  \begin{align}
  {\rm Ave}_w(\hat{\phi}) = \frac{\sum_{i=1}^{\hat{s}} \hat{\phi}_i (\gamma_i - \gamma_+)}{\sum_{i=1}^{\hat{s}}(\gamma_i - \gamma_+)}.
  \label{phihat}
  \end{align}
\end{enumerate}

%% file: section4.tex
\section{Numerical Experiments}
\label{sec4}
In this section, we examine how test accuracy behaves with respect to the proposed metric $\mathrm{Ave}_w(\hat{\phi})$ in~\eqref{phihat}.
We also compare the estimated thresholds obtained by the BEMA and Gaussian broadening methods to determine the one that is more appropriate using the proposed metric.
To examine the convergence accuracy of \eqref{phi} in the finite-sample setting, we compare the true value $\phi_i$ with its corresponding asymptotic limit $\hat{\phi_i}$ given in \eqref{phi}.
For the synthetic data, we generate a rank-2 matrix $W_{\rm signal}$ with $(\theta_1, \theta_2)=(3, 2)$, using randomly generated left and right orthogonal matrices.
In addition, the parameter $\sigma$ in $W_{\rm noise}$ is set to 1, and each entry in Table~\ref{tab:phi_by_size} is the average of results computed over 100 random matrices for $W_{\rm noise}$.
We confirm that $\hat{\phi}_i$ approximates $\phi_i$ well even in the finite-sample setting. 
\begin{table}[H]
\centering
\caption{The squared cosine similarity $\phi_i$ and its asymptotic limits $\hat{\phi}_i$ for $(\theta_1, \theta_2) = (3, 2)$ with $\sigma=1$}
\vspace{2mm}
\label{tab:phi_by_size}
\renewcommand{\arraystretch}{1.15}
\begin{tabular}{ccccc}
\toprule
\textbf{Matrix size} & $\phi_1$ & $\hat{\phi}_1$ & $\phi_2$ & $\hat{\phi}_2$ \\
\midrule
$100\times200$  & 0.940 & 0.939 & 0.859 & 0.860 \\
$250\times500$  & 0.941 & 0.940 & 0.861 & 0.862 \\
$500\times1000$ & 0.941 & 0.940 & 0.860 & 0.861 \\
\bottomrule
\end{tabular}
\end{table}
Next, we examine the relationship between the proposed metric $\mathrm{Ave}_w(\hat{\phi})$ and the test accuracy of trained DNNs.
Martin and Mahoney \cite{martin21a} pointed out that compared with smaller DNNs, the singular value distributions of larger DNNs deviate from the MP distribution and often exhibit heavy-tailed behavior.
The greater the classification difficulty, the more likely heavy tails are to appear, as reported in Meng and Yao \cite{Meng23}.
Thamm et al. \cite{thamm22} conducted a statistical test of the heavy-tailed hypothesis for DNN models. 
The experiments in this study used three models for which the heavy-tailed hypothesis was rejected in Thamm et al. \cite{thamm22}: a three-layer MLP, LeNet \citep{lecun02}, and AlexNet \citep{krizhevsky17}.
As the original LeNet architecture is too small for analysing the distribution of singular values, we modified the network by increasing the number of filters in the convolutional layers (Conv2D) and merged the three fully connected layers (FC) into a single large linear layer.
The detailed network architectures are provided in Appendix \ref{app3}.
In the same way as Thamm et al. \cite{thamm22}, we tested whether the singular values above $x_{\min}$ follow a power-law distribution $p(x) \propto x^{-\alpha_0}$ with tail index $\alpha_0$ for all FC layers of AlexNet, the largest of the three models trained on CIFAR-10.
As a result, the heavy-tailed hypothesis was rejected. 
All layers in the networks use the ReLU activation function, except for the output layer, which employs the softmax function. 
We trained all the models using the SGD from Glorot uniform initialization and normalized each RGB channel of the input images.
Batch size was set to $64$ for MLP and LeNet and to $128$ for AlexNet, fixing the learning rate at $0.01$. All models were trained for $30$ epochs.
 It should be noted that an excessive reduction in matrix dimensions should be avoided when reshaping convolutional kernels. This study employs the configuration of Zhang et al. \cite{zhang16}.
The parameters for BEMA and Gaussian broadening are $\alpha = 0.2$ and $a = 15$, respectively, with $\beta$ in \eqref{threshold} set to $0.1$. 
These values of $\alpha$ and $\beta$ are suggested as good choices for most settings in Ke et al. \cite{zheng21}.

\begin{figure}[H]
  \centering

  \begin{subfigure}[b]{0.49\textwidth}
    \centering
    \vspace{-1em}
    \includegraphics[width=\linewidth]{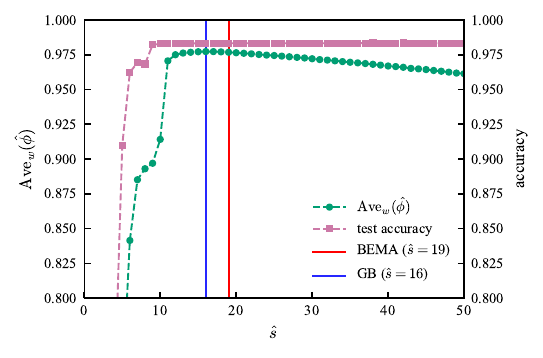}
    \vspace{-2.5em}
    \caption{Second layer of MLP on MNIST}
    \label{fig:mlp_mnist_3}
  \end{subfigure}\hfill
  \begin{subfigure}[b]{0.49\textwidth}
    \centering
    \vspace{-1em}
    \includegraphics[width=\linewidth]{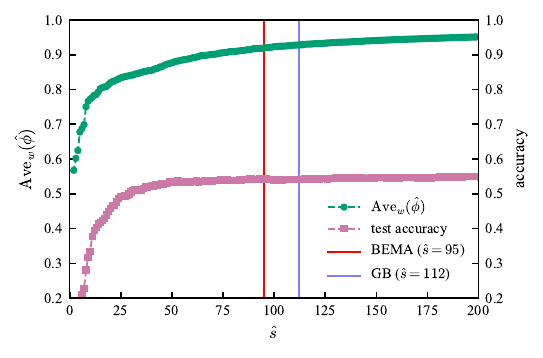}
    \vspace{-2.5em}
    \caption{Second layer of MLP on CIFAR-10}
    \label{fig:mlp_cifar_3}
  \end{subfigure}

  \vspace{1em}

  \begin{subfigure}[b]{0.49\textwidth}
    \centering
    \vspace{-1em}
    \includegraphics[width=\linewidth]{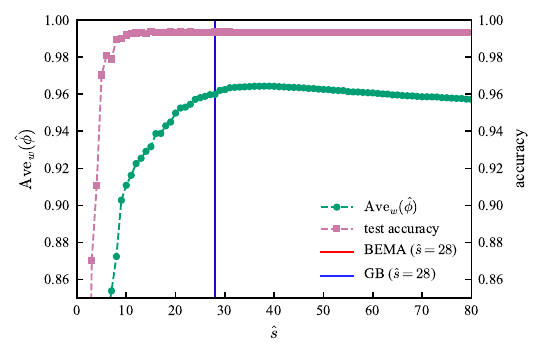}
    \vspace{-2.5em}
    \caption{Second convolutional layer of LeNet on MNIST}
    \label{fig:lenet_mnist_3}
  \end{subfigure}\hfill
  \begin{subfigure}[b]{0.49\textwidth}
    \centering
    \vspace{-1em}
    \includegraphics[width=\linewidth]{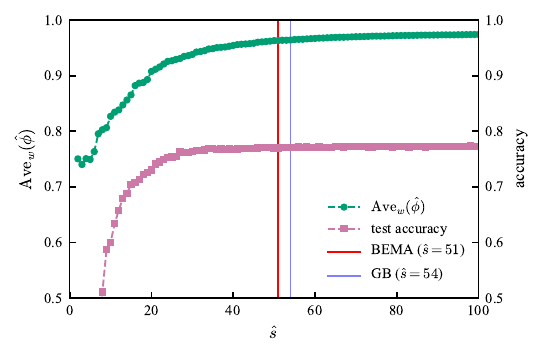}
    \vspace{-2.5em}
    \caption{Second convolutional layer of LeNet on CIFAR-10}
    \label{fig:lenet_cifar_3}
  \end{subfigure}

  \vspace{1em}

  \begin{subfigure}[b]{0.49\textwidth}
    \centering
    \vspace{-1em}
    \includegraphics[width=\linewidth]{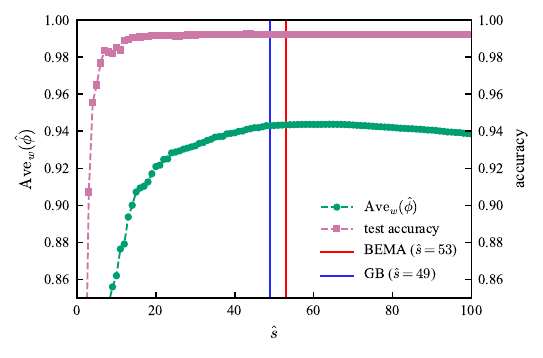}
    \vspace{-2.5em}
    \caption{Third convolutional layer of AlexNet on MNIST}
    \label{fig:alex_mnist_8}
  \end{subfigure}\hfill
  \begin{subfigure}[b]{0.49\textwidth}
    \centering
    \vspace{-1em}
    \includegraphics[width=\linewidth]{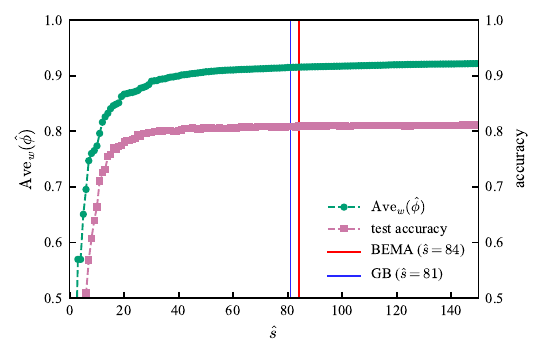}
    \vspace{-2.5em}
    \caption{Third convolutional layer of AlexNet on CIFAR-10}
    \label{fig:alex_cifar_8}
  \end{subfigure}

  \caption{Metric $\mathrm{Ave}_w(\hat{\phi})$ and test accuracy with respect to the estimated number of signal singular values ($\hat{s}$). Green circles (left $y$-axis) show $\mathrm{Ave}_w(\hat{\phi})$, and purple squares (right $y$-axis) show test accuracy obtained after keeping the top $\hat{s}$ singular values (others set to zero). Red and blue vertical lines indicate thresholds estimated by BEMA and Gaussian broadening, respectively.}
  \label{fig:cifar_acc}
\end{figure}

Figure~\ref{fig:cifar_acc} illustrates the relationship between \(\mathrm{Ave}_w(\hat{\phi})\) and test accuracy with increasing number of signal singular values $\hat{s}$. 
The left y-axis represents $\mathrm{Ave}_w(\hat{\phi})$, while the right y-axis indicates the test accuracy.
The singular values in the second linear layer of MLP, second Conv2D layer of LeNet, and third Conv2D layer of AlexNet, were reduced.
Test accuracy was observed to follow the behavior of the metric.
To quantify this relationship, we set $k$ to $20\%$ of the total number of singular values for each model and computed the correlation between $\mathrm{Ave}_w(\hat{\phi})$ and test accuracies determined over $\hat{s}=1,\dots,k$.
All reported values are the mean and standard deviation (SD) over 10 independent runs with different seeds.
On MNIST, the correlation coefficients were $0.918$ (SD: $0.008$), $0.859$ (SD: $0.041$), and $0.852$ (SD: $0.066$) for MLP, LeNet, and AlexNet, respectively, and on the CIFAR-10 dataset, they were $0.918$ (SD: $0.008$), $0.932$ (SD: $0.029$), and $0.919$ (SD: $0.032$).

Table \ref{tb:all} presents the values of the metric $\mathrm{Ave}_w(\hat{\phi})$ and $\hat{s}$. 
$\mathrm{Ave}_w(\hat{\phi})$ takes similar values regardless of whether BEMA or Gaussian-broadening.
Therefore, either approach yields no substantial differences in the low-rank approximations, and the resulting accuracy is expected to be similar.
\begin{table}[htbp]
  \centering
  \caption{$\mathrm{Ave}_w(\hat{\phi})$ and $\hat{s}$ (number of singular values exceeding the MP threshold) for MLP, LeNet, and AlexNet on MNIST and CIFAR-10, with thresholds estimated by BEMA and Gaussian broadening.}
  \label{tb:all}

  \vspace{0.75em}
  
  \textbf{MLP: first to third fully connected layers}
  \label{tb:mlp}

  \begin{tabularx}{\linewidth}{L{2.5em}  c cc cc | cc cc}
    \toprule
    & & \multicolumn{4}{c}{\textbf{MNIST}} & \multicolumn{4}{c}{\textbf{CIFAR-10}}\\
    \cmidrule(r){3-6}\cmidrule(l){7-10}
    \multirow{2}{*}{\textbf{Layer}} & \multirow{2}{*}{$\min(n,m)$}
      & \multicolumn{2}{c}{\textbf{BEMA}} & \multicolumn{2}{c}{\textbf{GB}}
      & \multicolumn{2}{c}{\textbf{BEMA}} & \multicolumn{2}{c}{\textbf{GB}}\\
    & & $\hat{s}$ & $\mathrm{Ave}_w(\hat{\phi})$ & $\hat{s}$ & $\mathrm{Ave}_w(\hat{\phi})$
      & $\hat{s}$ & $\mathrm{Ave}_w(\hat{\phi})$ & $\hat{s}$ & $\mathrm{Ave}_w(\hat{\phi})$\\
    \midrule
    FC\textsubscript{1} & 1024 & 51 & \textbf{0.909} & 47 & 0.906 & 226 & 0.969 & 260 & \textbf{0.971}\\
    FC\textsubscript{2} & 512  & 20 & 0.976 & 17 & \textbf{0.976} & 95  & 0.920 & 112 & \textbf{0.929}\\
    FC\textsubscript{3} & 350  & 10 &\textbf{0.974} & 10 & 0.970 & 61  & \textbf{0.890} & 61  & 0.889\\
    \bottomrule
  \end{tabularx}

\vspace{0.75em}

  \textbf{LeNet: second convolutional layer and first fully connected layer}
  \label{tb:lenet}

  \begin{tabularx}{\linewidth}{L{2.5em} c cc cc | cc cc}
    \toprule
    & & \multicolumn{4}{c}{\textbf{MNIST}} & \multicolumn{4}{c}{\textbf{CIFAR-10}}\\
    \cmidrule(r){3-6}\cmidrule(l){7-10}
    \multirow{2}{*}{\textbf{Layer}} & \multirow{2}{*}{$\min(n,m)$}
      & \multicolumn{2}{c}{\textbf{BEMA}} & \multicolumn{2}{c}{\textbf{GB}}
      & \multicolumn{2}{c}{\textbf{BEMA}} & \multicolumn{2}{c}{\textbf{GB}}\\
    & & $\hat{s}$ & $\mathrm{Ave}_w(\hat{\phi})$ & $\hat{s}$ & $\mathrm{Ave}_w(\hat{\phi})$
      & $\hat{s}$ & $\mathrm{Ave}_w(\hat{\phi})$ & $\hat{s}$ & $\mathrm{Ave}_w(\hat{\phi})$\\
    \midrule
    Conv2D\textsubscript{2} & 250 & 28 & \textbf{0.885} & 28 & 0.840 & 51 & 0.962 & 54 & \textbf{0.964} \\
    FC\textsubscript{1}     & 500 & 53 & \textbf{0.875} & 53 & 0.873 & 138  & \textbf{0.963} & 117 & 0.960 \\
    \bottomrule
  \end{tabularx}

  \vspace{0.75em}

  \textbf{AlexNet: second to fifth convolutional layers and first two fully connected layers}
  \label{tb:alexnet}

  \begin{tabularx}{\linewidth}{L{2.5em}  c cc cc | cc cc}
    \toprule
    & & \multicolumn{4}{c}{\textbf{MNIST}} & \multicolumn{4}{c}{\textbf{CIFAR-10}}\\
    \cmidrule(r){3-6}\cmidrule(l){7-10}
    \multirow{2}{*}{\textbf{Layer}} & \multirow{2}{*}{$\min(n,m)$}
      & \multicolumn{2}{c}{\textbf{BEMA}} & \multicolumn{2}{c}{\textbf{GB}}
      & \multicolumn{2}{c}{\textbf{BEMA}} & \multicolumn{2}{c}{\textbf{GB}}\\
    & & $\hat{s}$ & $\mathrm{Ave}_w(\hat{\phi})$ & $\hat{s}$ & $\mathrm{Ave}_w(\hat{\phi})$
      & $\hat{s}$ & $\mathrm{Ave}_w(\hat{\phi})$ & $\hat{s}$ & $\mathrm{Ave}_w(\hat{\phi})$\\
    \midrule
    Conv2D\textsubscript{2} & 320  & 35 & 0.969 & 33 & \textbf{0.969} & 59 & 0.965 & 61 & \textbf{0.966} \\
    Conv2D\textsubscript{3} & 576  & 42 & \textbf{0.943} & 39 & 0.943 & 85 & 0.903 & 89 & \textbf{0.906} \\
    Conv2D\textsubscript{4} & 768  & 54 & \textbf{0.944} & 50 & 0.943 & 84 & \textbf{0.915} & 81 & 0.915 \\
    Conv2D\textsubscript{5} & 768  & 47 & \textbf{0.935} & 43 & 0.933 & 49 & \textbf{0.909} & 44 & 0.908 \\
    FC\textsubscript{1}     & 1024 & 10 & \textbf{0.954} & 10 & 0.953 & 10 & \textbf{0.951} & 10 & 0.951 \\
    FC\textsubscript{2}     & 4096 & 11 & \textbf{0.956} & 11 & 0.953 & 10 & \textbf{0.942} & 10 & 0.941 \\
    \bottomrule
  \end{tabularx}
\end{table}

For the MNIST case, particularly in the linear layers, the values of \(\mathrm{Ave}_w(\hat{\phi})\) differ between BEMA and the Gaussian-broadening method even when both methods select the same \(\hat{s}\). 
This is because the metric evaluates the thresholds, and although the number of singular values exceeding the threshold is the same, the exact threshold values differ.
For the CIFAR-10 case, \(\hat{s}\) tends to be larger, and the singular-value distribution fits the MP distribution less well. Consequently, the estimated \(\hat{s}\) differ between the two methods, yet the resulting \(\mathrm{Ave}_w(\hat{\phi})\) values show no substantial difference.
For a LeNet model trained on CIFAR-10, the accuracies after applying low-rank approximation to all layers using BEMA and Gaussian broadening were both 76.3\%.
When the method with the higher metric value was selected and low-rank approximation was applied to each layer based on the proposed metric, the accuracy was 76.4\%, 
corresponding to a compression of 69.2\%, whereas the algorithm of \cite{idelbayev2020low} resulted in an accuracy of 72.6\% with a compression of 32.0\%. 
In this case, the RMT-based approach is also effective in terms of compression.

\begin{figure}[H]
  \centering

  \begin{subfigure}[b]{0.45\textwidth}
    \centering
    \includegraphics[width=\linewidth]{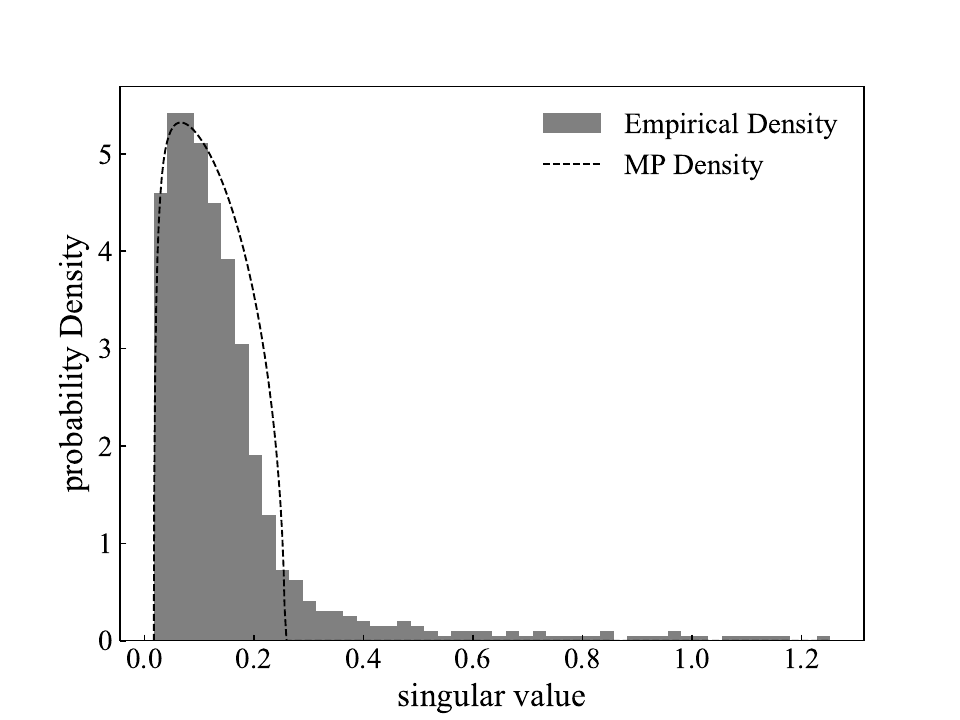}
    \caption{Batch size = 64}
    \label{fig:bt64}
  \end{subfigure}\hfill
  \begin{subfigure}[b]{0.45\textwidth}
    \centering
    \includegraphics[width=\linewidth]{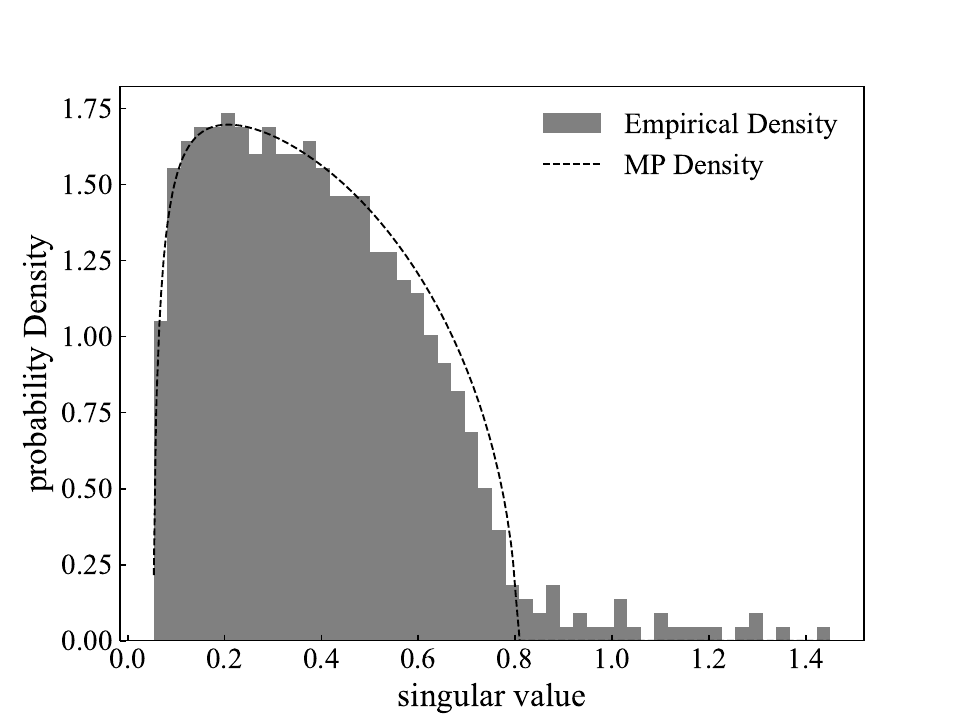}
    \caption{Batch size = 256}
    \label{fig:bt256}
  \end{subfigure}

  \caption{Singular value distribution of the FC\textsubscript{1} weight matrix in the MLP for different batch sizes. The dashed line represents the MP distribution estimated by BEMA, whereas the solid line indicates the threshold used to determine the number of singular values, $\hat{s}$, considered to represent the signal.}
  \label{fig:batch_svals}
\end{figure}

Finally, we investigated the behavior of the singular value distribution by varying the batch size, using the proposed metric. 
Figure~\ref{fig:batch_svals} shows the distribution of singular values of the FC1 weight matrix in MLP trained on MNIST for batch sizes of $64$ and $256$, with the learning rate fixed at $0.01$. 
The corresponding test accuracies are $98.43\%$ and $98.24\%$, respectively.
For a batch size of $64$, more singular values fall outside the support of the MP distribution than for a batch size of 256, and the largest singular values are substantially larger than the others.
A batch size of $256$ leaves only a few signal outliers and reduces the magnitudes of the largest singular values.
The metric \(\mathrm{Ave}_w(\hat{\phi})\) takes values of $0.950$ and $0.829$ for batch sizes of $64$ and $256$, respectively.
This indicates that excessively large batch sizes lead to performance degradation.
In practice, when the singular values that lie within the support of the MP distribution are removed, the corresponding test accuracy decreases from $98.3\%$ to $94.2\%$.

%% file: section5.tex
\section{Concluding remarks}
\label{sec5}
In this study, an evaluation metric was proposed for assessing the singular value thresholds $\gamma^2_+$ of the DNN weight matrices, based on the cosine similarity \eqref{phi} provided by Benaych-Georges and Nadakuditi \cite{georges12}.
We examined whether BEMA or Gaussian broadening provides a better approximation of the signal matrix.
In experimental results, the metric $\mathrm{Ave}_w(\hat{\phi})$ obtained from both methods was close, resulting in similar singular value thresholds and accuracies. 
However, the proposed metric allows for a quantitative determination of which low-rank approximation matrix is closer to the signal matrix.
This study considered only the case in which the model was trained using SGD. In future work, weight matrix $W$ optimized by methods other than SGD will be examined. We are currently working on an RMT-based low-rank approximation that takes this into consideration.

%% file: appendix.tex
\appendix
\label{appendix}
\section{BEMA algorithm}
\label{app1}
The BEMA algorithm proposed by Ke et al. \cite{zheng21} estimates the parameter $\sigma$ of the MP distribution.

\renewcommand{\thealgorithm}{}
\begin{algorithm}[H]
 \caption{Bulk Eigenvalue Matching Analysis}
\begin{algorithmic}[1]

\REQUIRE  Singular values of the weight matrix: \(\gamma_1, \dots, \gamma_m\),
Hyperparameters: \(\alpha \in (0, 1/2)\), \(\beta \in (0, 1)\)

\ENSURE  Estimated scale parameter of the MP distribution: \(\hat{\sigma}\)

\FOR{each \(\alpha m \leq k \leq (1 - \alpha) m\)}
  \STATE Let \( p_k \) be the upper \(k/m\) percentile point of the MP distribution with \(\sigma^2 = 1\),  
  such that \(\int_{p_k}^{1+\sqrt{q}} g(x)\,dx = \frac{k}{m}\)
\ENDFOR

\STATE Compute  
\[
\hat{\sigma} = \frac{\sum_{\alpha m \leq k \leq (1 - \alpha) m} p_k \gamma_k}
{\sum_{\alpha m \leq k \leq (1 - \alpha) m} p_k^2}
\]

\STATE Compute the upper $\beta$ percentile point of the Tracy--Widom distribution: \(t_{1 - \beta}\)
\end{algorithmic}
\end{algorithm}
The parameter \(\alpha\) determines the number of singular values used for estimating \(\sigma\). 
For example, if we set \(\alpha = 0.2\), the estimation of \(\sigma\) is performed using \(60\%\) of the singular values of the weight matrix \(W\), excluding the outermost \(20\%\) at both ends.  
The parameter \(\beta\) represents the significance level associated with the Tracy--Widom distribution.

\section{Gaussian broadening method}
\label{app2}
Gaussian broadening is a method for approximately estimating smooth continuous distributions from discrete data. 
It estimates a smooth distribution by superimposing Gaussian functions on each data point.
The smoothed empirical density is given by
\[
P(\gamma) \approx \frac{1}{m} \sum_{k=1}^m \frac{1}{\sqrt{2\pi \sigma_k^2}} \exp\left(-\frac{(\gamma - \gamma_k)^2}{2\sigma_k^2}\right),
\]
where the local standard deviation $\sigma_k$ is computed based on the spacing between neighboring singular values as
\(\sigma_k = (\gamma_{k+a} - \gamma_{k-a})/{2},\)
where the hyperparameter $a$ specifies the half-width of the window, corresponding to a total window size of $2a + 1$.
To fit the smoothed empirical singular value density $P(\gamma)$ to the density function of the MP distribution $g(\gamma)$ given in \eqref{mar}, we estimated the optimal parameter $\hat{\sigma}$ by solving the nonlinear least-squares problem.
\[
\hat{\sigma} = \arg\min_{\sigma} \sum_{i=1}^{m} \left[ P(\gamma_i) - g(\gamma_i) \right]^2.
\]

\section{Network architectures}
\label{app3}
\subsection*{\bf 3-layer MLP (MNIST / CIFAR-10)}
\vspace{1mm}
\begin{enumerate}
  \item Input image (MNIST: \(28 \times 28 = 784\), CIFAR-10: \(32 \times 32 \times 3 = 3072\)) is flattened into a 1D vector.
  \item Fully connected layer: input dimension to 1024 units.
  \item Fully connected layer: 1024 to 512 units.
  \item Fully connected layer: 512 to 512 units.
  \item Fully connected layer: 512 to 10 output logits.
\end{enumerate}
\subsection*{\bf LeNet (MNIST / CIFAR-10)}
\vspace{1mm}
\begin{enumerate}
  \item Input features (MNIST: $28 \times 28$, CIFAR-10: $32 \times 32 \times 3$) passed through a $5 \times 5$ convolution to 6 output channels.
  \item $2 \times 2$ max pooling with stride 2.
  \item $5 \times 5$ convolution with 16 output channels.
  \item $2 \times 2$ max pooling with stride 2.
  \item Fully connected layer from $256$ to 120 for MNIST, from $400$ to 120 for CIFAR-10.
  \item Fully connected layer from 120 to 84.
  \item Fully connected layer from 84 to output 10 logits.
\end{enumerate}
\subsection*{\bf AlexNet (MNIST / CIFAR-10)}
\begin{enumerate}
  \item Input features (MNIST: $28 \times 28$, CIFAR-10: $32 \times 32 \times 3$) passed through a $3\times3$ convolution to 96 output channels.
  \item $2 \times 2$ max pooling with stride 2.
  \item $3 \times 3$ convolution with 256 output channels.
  \item $2 \times 2$ max pooling with stride 2.
  \item $3 \times 3$ convolution with 384 output channels.
  \item $3 \times 3$ convolution with 384 output channels.
  \item $3 \times 3$ convolution with 256 output channels.
  \item $2 \times 2$ max pooling with stride 2.
  \item Flattened to a 4096 dimensional feature vector.
  \item Fully connected layer from 4096 to 1024.
  \item Fully connected layer from 1024 to 512.
  \item Fully connected layer from 512 to output 10 logits.
\end{enumerate}